\documentclass[10pt]{article}

\usepackage[numbers,square,sort,comma]{natbib}
\usepackage[preprint]{tmlr}
\setcitestyle{numbers,square,sort,comma}

\usepackage[hypertexnames=false]{hyperref}
\usepackage{url}
\usepackage{graphicx}
\usepackage{color}
\usepackage{listings}
\usepackage{xcolor}
\usepackage{threeparttable}
\usepackage{tabularx}
\usepackage{float}
\usepackage{makecell}
\usepackage{seqsplit}
\usepackage[noorphans]{quoting}
\usepackage{amsmath}
\usepackage{amssymb}
\usepackage{bm}
\usepackage{mathtools}
\usepackage{amsthm}
\usepackage{booktabs}
\usepackage{subcaption}
\usepackage{wrapfig}
\usepackage{caption}
\usepackage[most]{tcolorbox}
\usepackage{placeins}
\usepackage{dblfloatfix}

\newcommand{\code}[1]{\texttt{\seqsplit{#1}}}
\makeatletter
\renewcommand\paragraph{\@startsection{paragraph}{4}{\z@}%
  {1.25ex \@plus 1ex \@minus .2ex}%
  {0.75ex \@plus .2ex}%
  {\normalfont\normalsize\bfseries}}
\makeatother

\theoremstyle{plain}

\theoremstyle{definition}

\theoremstyle{remark}

\allowdisplaybreaks

\definecolor{codegreen}{rgb}{0,0.6,0}
\definecolor{codegray}{rgb}{0.5,0.5,0.5}
\definecolor{codepurple}{rgb}{0.58,0,0.82}
\definecolor{backcolour}{rgb}{1,1,1}

\title{Why Does Graph Learning Fail to Fully Benefit from a Text Teacher?}

\author{\name Fumiaki Kimino \email d2025fk@nii.ac.jp \\
  \addr SOKENDAI
  \AND
  \name Ryoma Sato \email rsato@nii.ac.jp \\
  \addr National Institute of Informatics, SOKENDAI
}

\begin{document}

\maketitle

\begin{abstract}
Graph neural networks (GNNs) are widely used across diverse real-world domains because they can effectively represent complex interactions and encode relationships among entities. We investigate a multimodal model that combines two complementary ideas: a self-supervised method that enables a GNN encoder pretrained on one dataset to operate directly on another dataset with a different node-feature dimensionality, without rebuilding the model or realigning the data; and an alternating optimization method that updates a language-model module in an E-step and a GNN module in an M-step, rather than jointly training a large language model and a GNN end to end on a large graph. Despite the expectation that these components would reinforce one another, the resulting model did not yield a sufficient improvement in predictive performance. This paper analyzes why. The experimental results suggest six factors: (1) an external anchor in the E-step has a strength--safety trade-off---a weak anchor has little effect, whereas an overly strong anchor can damage the graph representation; (2) the knowledge of the E-step teacher is not injected directly into the GCN embedding $Z$; (3) the representation space constructed in the M-step is not optimized for the same objective as the E-step teacher space, resulting in a compromise representation for target classification; (4) GCN propagation averages a node's own textual information with information from its neighbors; (5) cosine alignment does not guarantee axes that are discriminative for classification, so stronger geometric alignment with the E-step text anchor need not sufficiently improve the target decision boundary or classification performance; and (6) the force that preserves the source-side self-supervised geometry in the M-step conflicts with the force that moves the representation toward the E-step teacher. We support these observations through a staged set of experiments that varies the influence of the E-step.
\end{abstract}

\section{Introduction}

Graph neural networks (GNNs)~\cite{velickovic2018graph} are widely used in real-world applications because they can effectively represent complex interactions and encode relational structure among entities. Representative applications include scholarly networks~\cite{hu2020open}, review networks~\cite{wu2022diffnetpp}, and protein--protein interaction and protein-interface prediction~\cite{fout2017protein}. In node classification, knowledge learned from a source network can be transferred to predict node labels in a target network~\cite{dai2023adagcn,qiao2023sgda}. GNNs can also be applied to link prediction in the target network~\cite{kipf2016vgae}. 

Graph data are nevertheless highly complex, and a model trained on one graph dataset does not necessarily perform well on another~\cite{gui2022good}.

One model relevant to cross-domain transfer under incompatible feature formats is the Feature-Universal Graph pre-training model (FUG)~\cite{zhao2024fug}. FUG is a self-supervised graph pretraining model that learns the distribution within each feature dimension and can process node features of arbitrary formats. It enables a GNN encoder pretrained on one dataset to be applied directly to another dataset with a different node-feature dimensionality, without reconstructing the model or explicitly aligning the features across datasets.

For example, suppose that a review domain contains an information-rich source network in which many users have labels indicating their interests or ratings, as well as a newly formed target network without labels. Transferring useful knowledge from the source network may enable appropriate recommendations for unlabeled users in the target network~\cite{zhao2019ppgn,zhao2020catn}. If label or link prediction is possible even when the source and target networks belong to different domains, the range of graph data available for downstream applications can be substantially expanded~\cite{zhu2021transfer,huang2025enhancing}.

It is also important to broaden applicability by constructing multimodal models that combine graph structure and textual information. GLEM~\cite{zhao2023glem} addresses this problem through a variational expectation--maximization (EM) framework for efficiently learning from large text-attributed graphs. Instead of training a language model and a GNN jointly, GLEM alternates between an E-step that updates the language model using pseudo-labels predicted by the GNN and an M-step that updates the GNN using embeddings and pseudo-labels produced by the language model. The two modules can therefore be trained separately while still interacting and mutually reinforcing one another, which provides a mechanism for incorporating text into graph learning.

In scholarly networks, conventional node features often include publication year, number of authors, and venue. When the text of a paper is used, bag-of-words representations have also been common~\cite{yang2016planetoid}. Effective use of textual information such as paper abstracts would enable multimodal models that integrate graph structure with text, thereby broadening their applicability.

This paper considers cross-domain transfer in which the source and target networks differ in domain, while also seeking to incorporate text effectively into graph learning. We therefore combine FUG and GLEM, referring to the resulting family of models as FUG+GLEM. Using FUG as the graph component in the GLEM M-step allows the learned model to operate on a dataset whose node-feature dimensionality differs from that of the training dataset. Using a large text-attributed graph in the GLEM E-step is expected to improve the handling of text such as paper abstracts and make the model multimodal. We focus on the fundamental challenge of learning effective cross-domain node representations that can support applications such as node classification and link prediction.

At first glance, combining FUG and GLEM should outperform a FUG-only baseline. Surprisingly, however, our experiments showed no clear improvement over FUG alone. The purpose of this paper is therefore to determine why this outcome occurs and why an E-step teacher fails to sufficiently improve FUG in the M-step. We divide the investigation into two qualitatively different stages: one in which TextHead is trained to imitate the GCN (FUG) output, and another in which fixed external anchors---a raw text-hash anchor, and a frozen MPNet teacher---are used to compensate for weaknesses in the GCN. We compare the label-prediction performance of the resulting FUG+GLEM models with that of the FUG-only baseline. The experimental results suggest the following six factors. (1) The external anchor in the E-step exhibits a strength--safety trade-off: it has little effect when weak but can damage the graph representation when strong. (2) Knowledge from the E-step teacher is attenuated rather than being injected directly into the GCN embedding $Z$. (3) The M-step representation space is not constructed for the same objective as the E-step teacher space, and the resulting target representation becomes a compromise between the two. (4) GCN propagation averages each node's textual information with information from neighboring nodes. (5) Cosine alignment does not guarantee classification-relevant axes; although it can improve geometric alignment with the E-step text anchor, it need not sufficiently improve the decision boundary required for target classification. (6) The source-side self-supervised force that preserves the M-step geometry conflicts with the force that moves the representation toward the E-step teacher. These factors are suggested as explanations for why the E-step teacher fails to yield sufficient improvement in FUG during the M-step. We support this account through staged experiments that vary the influence of the E-step.

\section{Background}

We next describe FUG and GLEM, which provide the basis for the proposed method.

\subsection{FUG}

The Feature-Universal Graph Pre-training Model (FUG) is a self-supervised graph pretraining model that operates without class labels~\cite{zhao2024fug}. FUG learns the distribution within each feature dimension and generates a basis-transformation matrix $\mathbf{T}$ that maps arbitrary node-feature formats to a unified representation. Its theoretical motivation draws on work interpreting contrastive learning as generalized nonlinear PCA~\cite{shi2023tradeoff} and uses the concept of basis transformations from PCA~\cite{mackiewicz1993pca}. FUG then encodes both graph structure and node features to obtain graph representations~\cite{kipf2017semi}.

More specifically, FUG generates a basis-transformation vector for each feature dimension as follows~\cite{zhao2024fug}:

\begin{equation}
\label{eq:fug_basis}
\mathbf{t}_i = DE(\mathbf{X}_{:,i})
= \mathrm{Norm}\left[
\mathrm{MLP}\left(\hat{\mathbf{X}}_{:,i}^{\top}\right)
\right],
\end{equation}
where $\hat{\mathbf{X}} = \mathrm{Sample}(\mathbf{X})$ and $\mathrm{MLP}$ denotes a multilayer perceptron. The vector $\mathbf{X}_{:,i}$ is the $i$-th column of the feature matrix $\mathbf{X}$ and thus contains the values of the $i$-th feature over all nodes. A conventional neural network processes each row $\mathbf{X}_{v,:}$ of a feature matrix $\mathbf{X}\in\mathbb{R}^{n\times d}$, i.e., the feature vector of a node. In contrast, FUG feeds each column $\mathbf{X}_{:,i}$ to its Dimension Encoder. This column-wise processing encodes each feature dimension into a basis-transformation vector $\mathbf{t}_i$ according to its distribution, enabling datasets with different numbers of feature dimensions $d$ to be mapped to fixed-dimensional representations. Generating representations or parameters on a per-feature basis is related to Diet Networks~\cite{romero2017diet}; however, the objective of FUG is not feature selection but the mapping of heterogeneous feature formats into a shared representation space~\cite{zhao2024fug}.

FUG uses three loss terms, $L_{DE}$, $L_{\mathrm{RT\mbox{-}FUG}+}$, and $L_{\mathrm{RT\mbox{-}FUG}-}$, which we describe below.

Because the basis-transformation vectors should reflect differences among feature dimensions while preserving inter-dimensional correlations, FUG encourages the basis vectors to be distributed globally and uniformly over a hypersphere~\cite{zhao2024fug}. The loss for the Dimension Encoder is therefore defined as

\begin{equation}
\label{eq:fug_de}
L_{DE}(\mathbf{T})
=
\left\|
\frac{1}{d}
\sum_{i=1}^{d} \mathbf{t}_i
-
\mathbf{0}
\right\|_2^2 .
\end{equation}

To optimize relative relationships among nodes while bringing positive pairs closer together, FUG treats nodes connected by an edge as positive pairs~\cite{zhao2024fug}:

\begin{equation}
\label{eq:fug_positive}
L_{\mathrm{RT\mbox{-}FUG}+}
=
\frac{1}{|\mathbf{E}|}
\sum_{(v_i,v_j)\in \mathbf{E}}
\left\|
\mathbf{z}_i - \mathbf{z}_j
\right\|_2^2 .
\end{equation}

Here, $\mathbf{z}_j=\mathbf{Z}_{j,:}$ is the representation produced by the graph encoder for node $v_j$. Whether $\mathbf{z}_j$ is treated as a positive sample for $\mathbf{z}_i$ is determined not by how its representation is computed, but by whether an edge exists between $v_i$ and $v_j$, i.e., whether $(v_i,v_j)\in\mathbf{E}$.
This term brings together the representations of positive pairs, namely nodes connected by an edge. The construction assumes that adjacent nodes tend to have similar labels or semantic representations in homophilous graphs, consistent with prior work~\cite{zhang2022localized,lee2022augmentation}.

For negative pairs, accurately identifying negatives without prior knowledge is difficult and prone to sampling bias. FUG therefore avoids explicitly selecting negative pairs and instead constrains the representation globally~\cite{zhao2024fug}:

\begin{equation}
\label{eq:fug_negative}
L_{\mathrm{RT\mbox{-}FUG}-}
=
\left\|
\frac{1}{n}
\sum_{i=1}^{n}
\mathrm{Norm}(\mathbf{z}_i)
-
\mathbf{0}
\right\|_2^2 .
\end{equation}

The complete FUG objective jointly optimizes the three terms~\cite{zhao2024fug}:

\begin{equation}
\label{eq:fug_total}
L_{FUG}
=
\lambda_1 L_{\mathrm{RT\mbox{-}FUG}-}
+
\lambda_2 L_{\mathrm{RT\mbox{-}FUG}+}
+
\lambda_3 L_{DE}.
\end{equation}

\subsection{GLEM}

GLEM is a framework for integrating graph structure and language information on large text-attributed graphs. Rather than training a language model and a GNN jointly end to end, it uses a variational EM framework~\cite{neal1998view} to alternate between the two modules in an E-step and an M-step. The modules can therefore be trained separately while using one another's predictions as supervision~\cite{zhao2023glem}.

In the E-step, the language model learns from ground-truth labels $y_L$ on labeled nodes and from pseudo-labels predicted by the GNN on unlabeled nodes. In the M-step, the GNN is trained using node embeddings $\mathbf{h_V}$ and pseudo-labels $\hat{y}_U$ generated by the language model. More specifically, the GNN is optimized to predict $y_L$ on labeled nodes and $\hat{y}_U$ on unlabeled nodes. Repeating this process allows the modules to exchange predictions through pseudo-labels while retaining the computational and memory advantages needed for large graphs.

Let a text-attributed graph be
\begin{equation}
\label{eq:glem_graph}
G = (V, \mathbf{A}, \mathbf{s}_V),
\end{equation}
where $V$ is the node set, $\mathbf{A} \in \mathbb{R}^{|V| \times |V|}$ is the adjacency matrix, and $\mathbf{s}_V$ is the collection of textual attributes for all nodes. Each node $n\in V$ is associated with a sequential text feature, i.e., a sentence $s_n$.

GLEM is based on a pseudolikelihood variational framework. More specifically, it aims to maximize the log-likelihood of the observed node labels,
\begin{equation}
\label{eq:glem_loglikelihood}
\log p(y_L \mid \mathbf{s}_V, \mathbf{A}).
\end{equation}
Direct optimization is difficult because the labels $\mathbf{y}_U$ of the unlabeled nodes are unobserved. GLEM instead maximizes the following evidence lower bound (ELBO):
\begin{equation}
\label{eq:glem_elbo}
\log p(y_L \mid \mathbf{s}_V, \mathbf{A})
\geq
\mathbb{E}_{q(y_U \mid \mathbf{s}_U)}
\left[
\log p(y_L, y_U \mid \mathbf{s}_V, \mathbf{A})
-
\log q(y_U \mid \mathbf{s}_U)
\right].
\end{equation}
Here, $q(y_U \mid \mathbf{s}_U)$ is a variational distribution, and the inequality holds for any $q$.

The ELBO is maximized by alternating between optimization of $q$ in the E-step and optimization of $p$ in the M-step. The E-step updates $q$ to minimize the KL divergence between
\begin{equation}
\label{eq:glem_variational}
q(y_U \mid \mathbf{s}_U)
\end{equation}
and
\begin{equation}
\label{eq:glem_posterior}
p(y_U \mid \mathbf{s}_V, \mathbf{A}, y_L),
\end{equation}
thereby tightening the ELBO. In other words, the language-model distribution $q(y_U \mid \mathbf{s}_U)$, which predicts unlabeled-node labels from text alone, is brought closer to the posterior $p(y_U \mid \mathbf{s}_V, \mathbf{A}, y_L)$, which uses the graph structure, all textual attributes, and observed labels.

In the M-step, $q$ is fixed and $p$ is updated to maximize a pseudolikelihood defined as a product of node-wise conditional distributions~\cite{besag1975statistical}. The left-hand side of Eq.~\eqref{eq:glem_elbo} is the marginal log-likelihood of the observed labels $y_L$. The computationally difficult quantity is $p(y_L,y_U\mid\mathbf{s}_V,\mathbf{A})$ inside the expectation on the right-hand side. GLEM approximates it using a pseudolikelihood formed by the product of node-wise conditional distributions. Under this approximation, the GNN learns to predict each node's label from neighboring-node information and text-derived representations.

\section{Proposed Method}

\subsection{Problem Setting}

We consider cross-domain transfer learning in which a graph encoder trained on a source domain is applied to a target domain with a different feature distribution and graph structure~\cite{li2022ood}. Each graph comprises node features, an edge set, and labels used for evaluation. We use the trained graph model to generate target-domain embeddings and evaluate their classification performance with a linear probe.

\subsection{An EM-Like Framework for FUG+GLEM-ITT}

Initially, following the standard GLEM framework, we considered an EM-style training scheme in which the language model and graph encoder are updated using each other’s outputs as supervisory signals. However, our preliminary experiments suggested that feeding the current GCN representations back into TextHead would cause TextHead to do little more than reproduce those representations, making it difficult by design to provide the graph component with new textual information. The experimental results also supported this concern. In the proposed FUG+GLEM-ITT method, we consequently decouple TextHead from the current GCN output. TextHead is trained independently using a source-label classification loss on raw text-hash inputs, a self-supervised signal based on dropout augmentation, and an alignment loss to a raw-hash anchor obtained through a fixed projection of the raw text hash. Its representation is then used as an external anchor in the M-step. Because this design does not perform the bidirectional pseudo-label updates of standard GLEM, it is not a strict variational EM method. We instead regard it as EM-like learning inspired by GLEM's modular separation and alternating updates.

We replace the GNN module in GLEM with a FUG encoder and use the representation produced by a TextHead trained independently of the graph output as the TextHead anchor. We refer to this model as \emph{FUG+GLEM with an Independent Text Teacher} (FUG+GLEM-ITT). The name emphasizes that TextHead does not imitate the current GCN representation. Instead, it takes raw text-hash features as input and is optimized using a supervised source-label loss, a self-supervised consistency loss under dropout augmentation, and an alignment loss to a raw-hash anchor. The raw-hash anchor is a fixed reference representation obtained by mapping the raw text hash to the dimensionality of the GCN representation with a fixed random projection and then applying $\ell_2$ normalization. TextHead is therefore learned independently of both the graph encoder and its current output, and serves as a reference representation that constrains the graph encoder.

Training consists of independent TextHead pretraining, a self-supervised FUG warm-up, and alternating E-like and M-steps. In the E-like step, TextHead is not trained to imitate the current GCN representation $Z_{\mathrm{GCN}}$. It is instead optimized independently of the graph encoder using the raw text-hash features---the original word- and character-level feature-hashing representations of node text---together with the source-label classification loss, dropout-based self-supervision, and alignment to the fixed projection of the raw text hash. In the M-step, the FUG encoder is updated using the FUG self-supervised objective and alignment losses to four anchors: the TextHead anchor, the MLP-only anchor (the internal FUG representation before GCN propagation), the raw-hash anchor, and an external/random text anchor. The GCN representation obtained after an M-step is not used as a regression target or pseudo-label in the subsequent E-like step. The method is therefore an EM-like sequential learning procedure with an independent text-side teacher rather than a standard variational-EM implementation of GLEM. The numbers of pretraining epochs, iterations, and epochs per step are described in the experimental setup.

\begin{figure}[t]
    \centering
    \includegraphics[width=\linewidth]{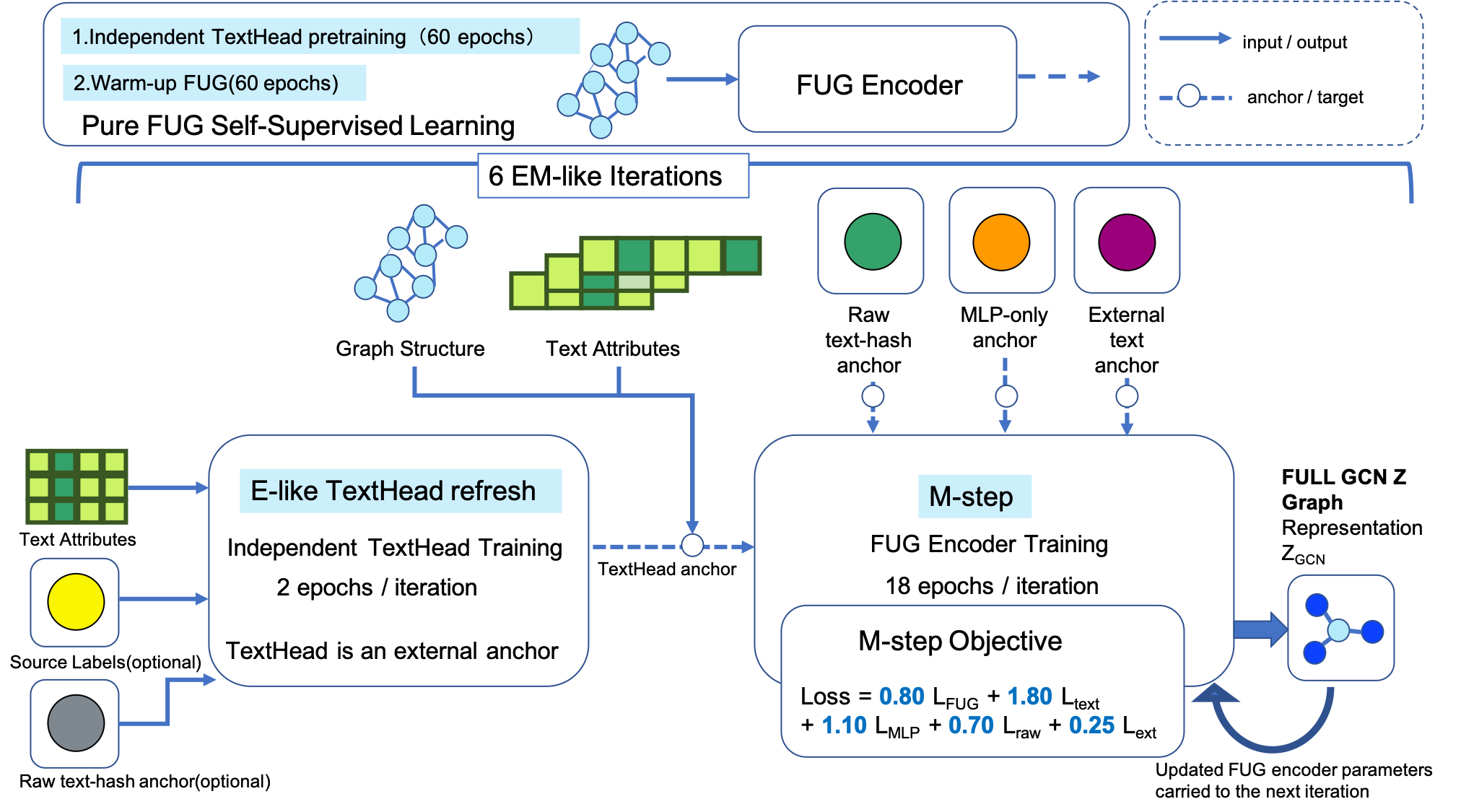}
    \caption{Overview of FUG+GLEM-ITT training. An independent TextHead that does not imitate the GCN output is first pretrained for 60 epochs using raw text-hash features, source labels, and dropout-based self-supervision. FUG is then warmed up for 60 epochs with its self-supervised objective, followed by six EM-like iterations. In the E-like step of each iteration, TextHead is refreshed for two epochs to produce an updated text anchor. In the M-step, the FUG encoder is updated for 18 epochs using the FUG self-supervised loss and auxiliary losses for the TextHead, MLP-only, raw text-hash, and external text anchors: $\mathcal{L}_{M}=0.80\mathcal{L}_{\mathrm{FUG}}+1.80\mathcal{L}_{\mathrm{text}}+1.10\mathcal{L}_{\mathrm{MLP}}+0.70\mathcal{L}_{\mathrm{raw}}+0.25\mathcal{L}_{\mathrm{ext}}$. The updated FUG parameters are carried forward to the next iteration.}
    \label{fig:fug_glem_itt_overview}
\end{figure}

Figure~\ref{fig:fug_glem_itt_overview} summarizes the training procedure.

\begin{enumerate}
\item[(1)] In the E-like step, TextHead is updated through a short refresh that uses the raw text-hash anchor, source labels, and a dropout-based text-side self-supervised signal, rather than the current graph-encoder output as a teacher. At the beginning of each iteration, this update produces a new text-anchor representation $\mathbf{A}_{\mathrm{text}}$.

\item[(2)] An \emph{anchor loss} is an auxiliary objective that moves the current GCN representation $\mathbf{Z}_{\mathrm{GCN}}$ toward a reference representation produced through a separate pathway. For a node $v$ in a set $S$, let $\bm{z}_v$ denote its GCN representation and $\bm{a}_v$ the corresponding anchor representation. We define the cosine anchor loss as

\begin{equation}
\label{eq:anchor_loss}
\mathcal{L}_{\mathrm{anchor}}
=
\mathcal{L}_{\cos}
\left(
\mathbf{Z}_{\mathrm{GCN}}, \mathbf{A}
\right)
=
\frac{1}{|S|}
\sum_{v \in S}
\left(
1-
\frac{
\bm{z}_v^{\top}\bm{a}_v
}{
|\bm{z}_v|_2
|\bm{a}_v|_2
}
\right).
\end{equation}
\end{enumerate}
We use the following four anchors.

\begin{enumerate}
\item[(i)] \textbf{TextHead anchor.}

This is the textual representation generated from the raw text-hash features $\mathbf{X}_{\mathrm{text}}$ by TextHead after its E-like update:
\begin{equation}
\label{eq:text_anchor}
\mathbf{A}_{\mathrm{text}}
=
\operatorname{sg}
\left(
T_{\phi}(\mathbf{X}_{\mathrm{text}})
\right),
\end{equation}
where $T_{\phi}$ is TextHead and $\operatorname{sg}(\cdot)$ denotes stop-gradient. The TextHead anchor loss is
\begin{equation}
\label{eq:text_anchor_loss}
\mathcal{L}_{\mathrm{text}}
=
\mathcal{L}_{\cos}
\left(
\mathbf{Z}_{\mathrm{GCN}},
\mathbf{A}_{\mathrm{text}}
\right).
\end{equation}
TextHead is fixed during the M-step, so only $\mathbf{Z}_{\mathrm{GCN}}$ is moved toward the TextHead representation.

\item[(ii)] \textbf{MLP-only anchor.}

We use the node representation $\mathbf{H}_{\mathrm{MLP}}$ obtained within the FUG encoder before GCN neighborhood propagation:
\begin{equation}
\label{eq:mlp_anchor}
\mathbf{A}_{\mathrm{MLP}}
=
\operatorname{sg}
\left(
\operatorname{Norm}
\left(
\mathbf{H}_{\mathrm{MLP}}
\right)
\right),
\end{equation}
where $\operatorname{Norm}(\cdot)$ denotes $\ell_2$ normalization. The MLP-only anchor loss is
\begin{equation}
\label{eq:mlp_anchor_loss}
\mathcal{L}_{\mathrm{MLP}}
=
\mathcal{L}_{\cos}
\left(
\mathbf{Z}_{\mathrm{GCN}},
\mathbf{A}_{\mathrm{MLP}}
\right).
\end{equation}
This loss encourages the post-propagation representation to retain information derived from the node's own features.

\item[(iii)] \textbf{Raw-hash anchor.}

The raw-hash anchor is a fixed reference representation obtained by mapping the raw text-hash features to the dimensionality of the GCN representation with a fixed random projection and then applying $\ell_2$ normalization:
\begin{equation}
\label{eq:raw_anchor}
\mathbf{A}_{\mathrm{raw}}
=
\operatorname{Norm}
\left(
\mathbf{X}_{\mathrm{text}}\mathbf{P}_{\mathrm{raw}}
\right),
\end{equation}
where $\mathbf{P}_{\mathrm{raw}}$ is a fixed random projection matrix that is not updated during training. The corresponding loss is
\begin{equation}
\label{eq:raw_anchor_loss}
\mathcal{L}_{\mathrm{raw}}
=
\mathcal{L}_{\cos}
\left(
\mathbf{Z}_{\mathrm{GCN}},
\mathbf{A}_{\mathrm{raw}}
\right).
\end{equation}
This term discourages the GCN representation from departing excessively from the structure of the original text features.

\item[(iv)] \textbf{External/random text anchor.}

The external/random text anchor is an auxiliary textual reference representation constructed independently of the GCN output. When an external text embedding $\mathbf{E}_{\mathrm{ext}}$ is available, we use its $\ell_2$-normalized form. Otherwise, we use a second fixed random projection of the raw text-hash features:
\begin{equation}
\label{eq:external_anchor}
\mathbf{A}_{\mathrm{ext}}
=
\begin{cases}
\operatorname{Norm}
\left(
\mathbf{E}_{\mathrm{ext}}
\right),
&
\text{if an external embedding is available},
\\[4pt]
\operatorname{Norm}
\left(
\mathbf{X}_{\mathrm{text}}\mathbf{P}_{\mathrm{ext}}
\right),
&
\text{otherwise}.
\end{cases}
\end{equation}
When this anchor and the GCN representation have different dimensions, the GCN representation is mapped into the same auxiliary space using a fixed projection matrix $\mathbf{P}_Z$:
\begin{equation}
\label{eq:projected_gcn}
\widetilde{\mathbf{Z}}_{\mathrm{GCN}}
=
\operatorname{Norm}
\left(
\mathbf{Z}_{\mathrm{GCN}}\mathbf{P}_Z
\right).
\end{equation}
The external/random text-anchor loss is
\begin{equation}
\label{eq:external_anchor_loss}
\mathcal{L}_{\mathrm{ext}}
=
\mathcal{L}_{\cos}
\left(
\widetilde{\mathbf{Z}}_{\mathrm{GCN}},
\mathbf{A}_{\mathrm{ext}}
\right).
\end{equation}
\end{enumerate}

These anchor losses are introduced to retain node-feature information---particularly text-derived information---in node representations that also reflect graph structure. Here, $Z_{\mathrm{GCN}}$ is the post-propagation node embedding generated by the FUG encoder from node features and graph structure. In the M-step, each cosine anchor loss is combined with the FUG self-supervised objective, and their weighted sum is minimized to move $Z_{\mathrm{GCN}}$ toward the four reference representations. Minimizing a term of the form $1-\cos(Z_{\mathrm{GCN}},A)$ moves the cosine similarity toward one and aligns the directions of the two representations. The FUG encoder is thereby updated to learn graph-structural information through its self-supervised loss while limiting the excessive loss of textual information during GCN propagation.

The M-step objective is
\begin{equation}
\label{eq:mstep_total}
\mathcal{L}_{M}
=
W_{\mathrm{FUG}}\mathcal{L}_{\mathrm{FUG}}
+
W_{\mathrm{text}}\mathcal{L}_{\mathrm{text}}
+
W_{\mathrm{MLP}}\mathcal{L}_{\mathrm{MLP}}
+
W_{\mathrm{raw}}\mathcal{L}_{\mathrm{raw}}
+
W_{\mathrm{ext}}\mathcal{L}_{\mathrm{ext}} .
\end{equation}

While retaining the FUG self-supervised objective, we assign relatively large weights to the independently learned TextHead anchor and the pre-GCN MLP-only anchor. We then align the post-propagation $Z_{\mathrm{GCN}}$ with these representations using cosine losses. This design limits excessive divergence from both the text-derived representation and the pre-GCN representation, constraining GCN propagation from excessively damaging text-derived semantic information.

\subsection{Representations Used for Evaluation}

We evaluate the post-propagation Full GCN representation ($\mathbf{Z}$), the Raw Text Hash representation, the Raw MPNet representation, and late-fusion ensembles formed by concatenating these representations along the feature dimension~\cite{weinberger2009featurehashing,song2020mpnet,baltrusaitis2019multimodal}. The standard ensemble concatenates Full GCN Z with Raw Text Hash, $[\mathbf{Z}_{\mathrm{GCN}};\mathbf{X}_{\mathrm{hash}}]$, and is evaluated using a linear probe under the same train/validation/test split. For Exp4, we additionally evaluate $[\mathbf{Z}_{\mathrm{GCN}};\mathbf{X}_{\mathrm{MPNet}}]$, which we call the GCN+MPNet ensemble~\cite{song2020mpnet,baltrusaitis2019multimodal}. Raw MPNet is obtained by directly encoding each node's text with MPNet, without applying the transformations used by the graph model. Because Exp4 uses MPNet as an external teacher, Raw MPNet provides a reference that separates the quality of the teacher itself from the effectiveness with which its information is transferred to Full GCN $Z$. Full GCN $Z$ is the final graph-encoder output and is the primary representation of interest. Raw Text Hash and Raw MPNet provide text-only reference representations that do not pass through the GCN.

\section{Experimental Setup and Main Result}

\subsection{Settings}

We use the Digital Music category of the Amazon product data as the source domain and OpenAlex as the target domain. The Amazon product data constitute a large review dataset containing reviews, ratings, product metadata, and product relations; we use the Digital Music category as the source review network~\cite{ni2019amazon,mcauley2015image}. OpenAlex is an open scholarly knowledge graph containing papers, authors, institutions, publication venues, concepts, and citation relationships; we use its paper network as the target domain~\cite{priem2022openalex}.

\subsection{Main Experimental Result}

FUG+GLEM-ITT did not clearly outperform the FUG-only baseline. For our primary metric, the standard-probe accuracy of Full GCN $Z$ on OpenAlex, FUG-only achieved $0.7459$, whereas FUG+GLEM-ITT with an external TextHead anchor achieved $0.7480$, an improvement of only approximately $+0.21$ percentage points. Under the balanced probe, the balanced accuracy of Full GCN $Z$ was $0.6016$ for FUG-only and $0.6001$ for FUG+GLEM, representing a slight decrease (Table~\ref{tab:fug_baseline_ITT_results}). These results do not support the hypothesis that FUG+GLEM would reliably outperform FUG by at least $1\%$. They instead motivate an analysis of why the E-step teacher failed to sufficiently improve FUG in the M-step.

\begin{table}[t]
\centering
\caption{Cross-domain transfer performance between the FUG baseline and FUG+GLEM-ITT.}
\label{tab:fug_baseline_ITT_results}
\begin{threeparttable}
\small
\setlength{\tabcolsep}{3.5pt}
\begin{tabularx}{\textwidth}{lXcccc}
\toprule
Experiment & Method & \makecell{Full GCN \\ ($Z$)} & \makecell{Raw Text \\ Hash} & \makecell{Raw \\ MPNet} & Ensemble \\
\midrule
Baseline & FUG-only & 0.7459 & 0.7623 & -- & 0.7702 \\
FUG+GLEM & FUG+GLEM-ITT & 0.7480 & 0.7623 & -- & 0.7717 \\
\bottomrule
\end{tabularx}
\begin{tablenotes}
\footnotesize
\item[] All values are standard-probe test accuracies on OpenAlex. The Ensemble column denotes the concatenation of Full GCN ($Z$) and Raw Text Hash.
\end{tablenotes}
\end{threeparttable}
\end{table}

\section{Analysis}

\subsection{Additional Experiments}

To investigate why FUG+GLEM-ITT failed to outperform the FUG baseline, we introduce Exp2, Exp3, and Exp4 and analyze the mechanism through their results.

\subsection{Design of the External E-Step Anchor}

Across Exp2, Exp3, and Exp4, we progressively vary the externality and semantic strength of the teacher information. An external anchor is a reference representation supplied to the graph encoder separately from the FUG self-supervised objective, constraining the output in a particular direction. \emph{Externality} denotes the degree to which the anchor is generated independently of the current graph-encoder output. \emph{Semantic strength} denotes the extent to which the anchor captures semantic content beyond surface-level lexical information.

Exp2 uses a self-copy TextHead that imitates the current GCN representation $Z_{\mathrm{GCN}}$~\cite{chen2021simsiam}. The current $Z_{\mathrm{GCN}}$ is fixed as a teacher representation, and TextHead is trained to approach it. Because the teacher information is derived from the graph encoder's own output, its externality and newly introduced semantic information are weakest in this setting.

Exp3 introduces an external anchor based on the raw text hash~\cite{weinberger2009featurehashing}. The raw text hash is a fixed-dimensional representation of word- and character-level features generated from node text through feature hashing. Because it does not depend on the current GCN output, it is more external than the Exp2 teacher. Its semantic strength is nevertheless limited because it primarily represents lexical and character-string occurrence patterns.

Exp4 uses a pretrained, frozen MPNet as an external teacher and supplies the resulting text embedding to the graph encoder as an anchor~\cite{song2020mpnet}. This representation is independent of the current GCN output and contains contextual semantic information; we therefore treat it as having the greatest externality and semantic strength among Exp2--Exp4. The experiments thus progress from a graph-encoder-derived representation, through surface-level textual features, to a semantic representation from a pretrained language model.

\subsection{M-Step Objectives}

In Exp2, the FUG self-supervised objective is augmented with a teacher-derived distillation loss~\cite{sanh2019distilbert}. The current FUG/GCN encoder is frozen to obtain
\begin{equation}
\label{eq:exp2_frozen}
\mathbf{Z}_{\mathrm{frozen}} = \mathrm{sg}(\mathbf{Z}_{\mathrm{GCN}}),
\end{equation}
and TextHead $T_{\phi}$ is trained with
\begin{equation}
\label{eq:exp2_estep}
\mathcal{L}_{E}^{\mathrm{Exp2}} = \mathcal{L}_{\cos} \left( T_{\phi}(\mathbf{X}_{\mathrm{text}}), \mathbf{Z}_{\mathrm{frozen}} \right).
\end{equation}
Here, $\mathrm{sg}(\cdot)$ denotes stop-gradient~\cite{grill2020byol}. The Exp2 M-step objective is
\begin{equation}
\label{eq:exp2_mstep}
\mathcal{L}_{M}^{\mathrm{Exp2}} = \mathcal{L}_{\mathrm{FUG}} + \lambda_{\mathrm{distill}} \mathcal{L}_{\cos} \left( \mathbf{Z}_{\mathrm{GCN}}, \mathrm{sg}\left(T_{\phi}(\mathbf{X}_{\mathrm{text}})\right) \right).
\end{equation}
Exp3 uses the raw text-hash anchor~\cite{weinberger2009featurehashing} and the MLP-only anchor~\cite{grill2020byol}:
\begin{equation}
\label{eq:exp3_mstep}
\mathcal{L}_{M}^{\mathrm{Exp3}} = \mathcal{L}_{\mathrm{FUG}} + \lambda_{\mathrm{raw}} \mathcal{L}_{\cos} \left( \mathbf{Z}_{\mathrm{GCN}}, \mathbf{A}_{\mathrm{raw}} \right) + \lambda_{\mathrm{MLP}} \mathcal{L}_{\cos} \left( \mathbf{Z}_{\mathrm{GCN}}, \mathrm{sg}\left(\mathbf{A}_{\mathrm{MLP}}\right) \right).
\end{equation}
Exp4 combines an MPNet-teacher cosine anchor loss~\cite{song2020mpnet,reimers2019sbert},
\begin{equation}
\label{eq:analysis_anchor_loss}
\mathcal{L}_{\mathrm{anchor}} = \mathcal{L}_{\cos} \left( \mathbf{Z}_{\mathrm{GCN}}, \mathbf{A} \right) = \frac{1}{|S|} \sum_{v \in S} \left( 1 - \frac{\bm{z}_v^{\top}\bm{a}_v}{|\bm{z}_v|_2 |\bm{a}_v|_2} \right),
\end{equation}
with the MLP-only anchor~\cite{chen2021simsiam}. Its complete objective is
\begin{equation}
\label{eq:exp4_mstep}
\mathcal{L}_{M}^{\mathrm{Exp4}} = \mathcal{L}_{\mathrm{FUG}} + \lambda_{\mathrm{MPNet}} \mathcal{L}_{\cos} \left( \mathbf{Z}_{\mathrm{GCN}}, \mathbf{A}_{\mathrm{MPNet}} \right) + \lambda_{\mathrm{MLP}} \mathcal{L}_{\cos} \left( \mathbf{Z}_{\mathrm{GCN}}, \mathrm{sg}\left(\mathbf{A}_{\mathrm{MLP}}\right) \right).
\end{equation}
Here, $\mathcal{L}_{\mathrm{FUG}}$ is the FUG self-supervised objective; $\mathcal{L}_{\cos}$ is a cosine loss~\cite{chen2021simsiam,reimers2019sbert}; $\mathbf{Z}_{\mathrm{GCN}}$ is the post-propagation graph representation~\cite{kipf2017semi}; $T_{\phi}(\mathbf{X}_{\mathrm{text}})$ is the text-derived TextHead representation; $\mathbf{A}_{\mathrm{raw}}$, $\mathbf{A}_{\mathrm{MPNet}}$, and $\mathbf{A}_{\mathrm{MLP}}$ are the raw text-hash, frozen-MPNet, and MLP-only anchors, respectively; and $\mathrm{sg}(\cdot)$ denotes stop-gradient~\cite{grill2020byol,chen2021simsiam}. The FUG loss preserves source-graph structural geometry, whereas the anchor loss moves the graph representation toward a text-based teacher representation. We vary the distillation and anchor weights to control the influence of the teacher~\cite{hinton2015distilling}.

\subsection{Findings}

Why did FUG+GLEM-ITT fail to outperform the baseline despite the presence of a Teacher? What is the underlying mechanism? Our investigation of why FUG+GLEM-ITT failed to outperform the FUG baseline suggests a broader mechanism: external anchors in the FUG+GLEM-ITT EM-like procedure exhibit a trade-off between strength and safety. In other words, a weak external anchor has little effect, whereas a strong anchor can damage the graph representation. Through four staged experiments, we analyze why the E-step could not sufficiently improve the self-supervised FUG model in the M-step and argue that the experimental results suggest this trade-off.

The stages can be summarized as follows: Exp1 is the FUG-only baseline; Exp2 uses TextHead to imitate the GCN (FUG) output; Exp3 introduces a fixed raw text-hash anchor; and Exp4 introduces frozen MPNet as a fixed teacher. The maximum number of graph-encoder update epochs executed to generate candidate models is matched across the experiments. FUG-only is trained for 168 epochs, while the FUG+GLEM settings use 60 epochs of FUG warm-up followed by six iterations with 18 M-step epochs each, for a total of $60+6\times18=168$ epochs. The limited improvement therefore cannot be attributed solely to an insufficient maximum number of graph-encoder update epochs; the results also suggest a bottleneck in transferring teacher or anchor information into GCN $Z$.

The experiments also introduce staged differences in anchor strength. Exp2 uses \code{EXP2\_W\_DISTILL}$=0.80$ for the self-copy TextHead that imitates GCN pseudo-$z$. Exp3 uses \code{EXP3\_W\_RAW}$=1.20$ for the raw-hash anchor and \code{EXP3\_W\_MLP}$=0.80$ for the MLP-only anchor. Exp4 uses \code{EXP4\_W\_MPNET}$=1.00$ for the MPNet anchor and \code{EXP4\_W\_MLP}$=0.30$ for the MLP-only anchor; its raw auxiliary anchor is disabled by default. Thus, Exp2 represents imitation of the GCN itself, Exp3 uses an external raw text-hash anchor, and Exp4 uses a semantically stronger MPNet-based anchor. These settings progressively vary the externality and semantic strength of the anchors. Table~\ref{tab:cross_domain_results} reports cross-domain transfer performance across the four stages.

\subsection{Bottlenecks in Transferring External-Teacher Knowledge to GCN Representations}

For the OpenAlex evaluation, we scan the first 20,000 records in the JSONL file and retain 6,984 nodes that contain both text and labels and belong to one of the 20 most frequent classes. Evaluation uses the \code{concept\_or\_type} mode: an OpenAlex concept is used when available, and a document-type label is used only when no concept is present. Candidate regularization parameters for the linear probe are $C\in\{0.01,0.1,1.0,10.0,50.0\}$. All OpenAlex accuracies are therefore compared under the same data split and probe conditions.

The graph-side mechanism in FUG+GLEM-ITT can be summarized as follows. Teacher knowledge is first projected into a fixed-dimensional anchor, but this anchor is not directly fed into the FUG MLP-only representation. Instead, the FUG encoder is updated through a cosine-alignment loss between Full GCN $Z$ and the anchor. Under the updated encoder, node features are transformed into an MLP-only representation and subsequently propagated through a two-layer GCN to produce Full GCN $Z$. External-teacher knowledge is therefore not injected directly into GCN $Z$; it affects the final representation indirectly through anchor projection and alignment-based encoder updates. We hypothesize that the external anchor fails to sufficiently improve FUG because target-relevant semantic information in the teacher is compressed, deformed to accommodate the source-graph geometry, and further diluted by neighborhood averaging, leaving insufficient information for the target decision boundary.

\paragraph{Teacher knowledge is not injected directly into GCN $Z$.}

Consider Exp4, which uses an MPNet teacher~\cite{song2020mpnet}. Raw MPNet is highly effective on OpenAlex, achieving a standard accuracy of $0.7888$. In contrast, Full GCN $Z$ reaches only $0.7437$ when MPNet is incorporated as a teacher in the EM procedure. The GCN+MPNet ensemble, which combines GCN $Z$ and MPNet only at the final stage, achieves $0.7845$ (see the Raw MPNet, Full GCN $Z$, and Ensemble columns of the Exp4 row in Table~\ref{tab:cross_domain_results}). These results indicate that MPNet itself contains target-relevant semantic information, but performance decreases when that information is transferred to GCN $Z$. The bottleneck therefore appears to lie in the transfer pathway from the MPNet semantic representation to GCN $Z$, rather than in the quality of MPNet itself.

\paragraph{The MPNet and raw-text semantic spaces are projected into the GCN-$Z$ space.}

Text-derived information reaches GCN $Z$ only after undergoing dimensional transformation. In Exp4, the original 768-dimensional MPNet embeddings are mapped to a 1,024-dimensional anchor and influence GCN $Z$ through cosine alignment rather than being fed directly into the GCN. Similarly, the 2,048-dimensional Raw Text Hash representations are mapped to the 1,024-dimensional hidden space used by FUG+GLEM-ITT. These transformations may distort the semantic geometry of MPNet and compress the lexical information contained in Raw Text Hash. Teacher knowledge is therefore introduced as a directional constraint rather than transferred as an intact representation space. In this process, the projection and subsequent alignment do not necessarily preserve the fine-grained directions useful for OpenAlex classification. This interpretation is consistent with the observed performance gap between the text-only representations and Full GCN $Z$. Further details on the dimensional transformations, their implications for classification, and their relation to representation-level knowledge transfer are provided in Appendix~\ref{app:text_to_gcn_projection}.

\paragraph{The FUG representation space is not designed for the same objective as the teacher representation space}
The Full GCN $Z$ representation in FUG is not constructed directly for supervised classification; rather, it is learned through self-supervised losses to reflect neighborhood structure and stabilize the overall representation. By contrast, MPNet and Raw Text Hash encode semantic and lexical similarities, respectively, and are directly useful for OpenAlex concept classification. Thus, the objectives underlying the teacher space, the FUG/GCN $Z$ space, and the OpenAlex target space are not necessarily aligned. Even when a teacher anchor is introduced, GCN $Z$ does not exclusively incorporate directions that are useful for the teacher's classification performance. Instead, it becomes a compromise representation shaped by the FUG self-supervised losses, GCN propagation, and constraints imposed by the source graph. Indeed, in Exp4, alignment with MPNet improved, whereas the OpenAlex balanced accuracy of Full GCN $Z$ remained at $0.7437$. Further details are provided in Appendix~\ref{app:Mismatch between the Objectives of the Teacher and FUG Representation Spaces}.

\paragraph{Potential dilution of node-specific textual information through GCN propagation}
One possible mechanism that may explain these results is the mixing of a node's own textual information with neighborhood information through GCN propagation. Conceptually, the FUG pipeline proceeds from raw text features through the Dimension Encoder or an MLP-only representation, followed by GCN propagation to obtain Full GCN $Z$. The MLP-only representation is a text-derived representation obtained before GCN propagation, whereas Full GCN $Z$ incorporates neighborhood information through the subsequent GCN layers.

Neighborhood aggregation can be effective when edges predominantly connect nodes with the same label. In a cross-domain setting, however, the neighborhood-smoothing behavior learned from the source graph is not necessarily useful for the target graph. In our setting, the source domain is Amazon Music and the target domain is OpenAlex, whose edges encode different types of relationships. Consequently, even if the teacher preserves node-specific semantic information, this discriminative information may be mixed with neighborhood information that is not necessarily useful for target classification and may therefore not be fully exploited after GCN propagation.

Indeed, in Exp4 and the FUG+GLEM-ITT-related experiments, Full GCN $Z$ underperformed Raw Text Hash in terms of balanced accuracy, and its improvement over the pre-propagation representations was small or negative in some settings. This pattern is consistent with the possibility that GCN propagation dilutes node-specific discriminative information derived from text. However, this observation is limited to balanced accuracy and does not establish a general performance degradation across all evaluation metrics or directly demonstrate the proposed mechanism. Detailed comparisons and limitations of this interpretation are provided in Appendix~\ref{app:Comparison of Balanced Accuracy}.

\paragraph{Cosine alignment does not guarantee classification-effective axes.}

Alignment metrics improve in Exp3 and Exp4. Table~\ref{tab:alignment_results} summarizes their changes together with the OpenAlex performance of Full GCN $Z$. In Exp3, introducing the raw text-hash anchor increases both \code{ood\_cos} and \code{raw\_cos}, but Full GCN $Z$ does not outperform FUG-only. In Exp4, \code{mpnet\_cos} increases substantially, but this increase does not translate into better Full GCN-$Z$ performance.

FUG+GLEM-ITT likewise learns to improve alignment with the TextHead anchor. The M-step value \code{text\_anc}, i.e., $1.0-(\mathbf{z_n}\cdot\mathrm{anchor})$, decreases over the EM-like iterations. The values of \code{mlp\_anc} and \code{raw\_anc} also decrease, demonstrating improved geometric alignment between the GCN representation and the anchors. Nevertheless, the resulting improvement in Full GCN-$Z$ performance on OpenAlex is limited (Table~\ref{tab:representation_performance}). Geometric alignment with a text anchor can therefore improve without sufficiently improving the target-task decision boundary or classification performance.

The reason is that a cosine loss moves each node's GCN representation toward the direction of its teacher vector but does not directly enforce the conditions required for classification. Improved classification requires, at a minimum, (i) wider inter-class margins, (ii) compact within-class representations, (iii) preservation of minority classes, and (iv) retention of axes associated with target labels. Cosine alignment directly optimizes none of these conditions. Even if every node moves somewhat closer to the teacher, performance will not improve unless inter-class distances increase. Similarly, global alignment with the teacher will not improve balanced accuracy or macro-F1 if GCN propagation removes the fine-grained axes separating minority classes. A teacher-aligned representation is therefore not necessarily the same as a representation in which target labels are easy to separate.

\paragraph{Potential tension between preserving the source-side FUG geometry and aligning with the teacher}
The M-step does not optimize the teacher anchor alone; it also retains the FUG SSL loss. In Exp4, the M-step is likewise constructed as a combination of FUG SSL and a frozen MPNet semantic anchor. Consequently, during training, GCN $Z$ is simultaneously subject to the objective of preserving the self-supervised representation learned from the source graph and the objective of approaching the text semantic anchor. However, the graph structure of Amazon Music, the textual semantic space of MPNet, and the structure required for OpenAlex concept classification are not necessarily aligned. The final GCN $Z$ is therefore expected to represent a compromise between the FUG SSL and teacher-alignment objectives.

This possibility is also reflected in the candidate-model selection results. In FUG+GLEM-ITT, em6 was selected because it achieved the highest composite score, although its source \code{valid\_bacc} of $0.4766$ was numerically slightly lower than the warmup value of $0.4788$. Conversely, SWA achieved \code{valid\_bacc}$=0.4789$ but was not selected because its composite score was lower than that of em6. These results suggest that, under the current configuration, preserving source validation performance and improving alignment with multiple anchors are not necessarily fully consistent objectives. However, the performance differences are small, and these results alone do not establish a causal conflict between the two objectives. Further details are provided in Appendix~\ref{app:Trade-off Between FUG SSL and Teacher Alignment}.

\section{Conclusion}

This paper investigated why FUG+GLEM-ITT failed to outperform the FUG baseline by designing and conducting additional experiments and analyzing their results. The analysis provides strong evidence regarding the underlying mechanism.

Taken together, the six findings presented in the \textit{Findings} subsection indicate that merely introducing a strong text-based Teacher and aligning the outputs of the graph encoder with the Teacher representations does not, by itself, guarantee an improvement in graph representation learning performance. Improving graph learning with text requires careful design not only of the teacher representation, but also of the pathway by which teacher knowledge is transferred into the graph encoder, the mechanism that preserves node-specific textual information after GCN propagation, and the objective that aligns the learned representation with the target-task decision boundary. The success of knowledge transfer should also be assessed with measures of whether discriminative information actually reaches the GCN representation, rather than with cosine alignment alone.

The implications extend beyond this particular FUG+GLEM-ITT implementation and provide broader design guidance for methods that seek to improve graph learning with textual information.

\newpage
\bibliographystyle{unsrt}
\bibliography{main}

\begin{thebibliography}{10}

\bibitem{velickovic2018graph}
Petar Veli{\v{c}}kovi{\'c}, Guillem Cucurull, Arantxa Casanova, Adriana Romero, Pietro Li{\`o}, and Yoshua Bengio.
\newblock Graph attention networks.
\newblock In {\em Proceedings of the 6th International Conference on Learning Representations (ICLR 2018)}, 2018.

\bibitem{hu2020open}
Weihua Hu, Matthias Fey, Marinka Zitnik, Yuxiao Dong, Hongyu Ren, Bowen Liu, Michele Catasta, and Jure Leskovec.
\newblock Open graph benchmark: Datasets for machine learning on graphs.
\newblock In {\em Advances in Neural Information Processing Systems 33 (NeurIPS 2020)}, pages 22118--22133, 2020.

\bibitem{wu2022diffnetpp}
Le~Wu, Junwei Li, Peijie Sun, Richang Hong, Yong Ge, and Meng Wang.
\newblock Diffnet++: A neural influence and interest diffusion network for social recommendation.
\newblock {\em IEEE Transactions on Knowledge and Data Engineering}, 34(10):4753--4766, 2022.

\bibitem{fout2017protein}
Alex Fout, Jonathon Byrd, Basir Shariat, and Asa Ben-Hur.
\newblock Protein interface prediction using graph convolutional networks.
\newblock In {\em Advances in Neural Information Processing Systems 30 (NeurIPS 2017)}, 2017.

\bibitem{dai2023adagcn}
Quanyu Dai, Xiao-Ming Wu, Jiaren Xiao, Xiao Shen, and Dan Wang.
\newblock Graph transfer learning via adversarial domain adaptation with graph convolution.
\newblock {\em IEEE Transactions on Knowledge and Data Engineering}, 35(5):4908--4922, 2023.

\bibitem{qiao2023sgda}
Ziyue Qiao, Xiao Luo, Meng Xiao, Hao Dong, Yuanchun Zhou, and Hui Xiong.
\newblock Semi-supervised domain adaptation in graph transfer learning.
\newblock In {\em Proceedings of the Thirty-Second International Joint Conference on Artificial Intelligence (IJCAI 2023)}, 2023.

\bibitem{kipf2016vgae}
Thomas~N. Kipf and Max Welling.
\newblock Variational graph auto-encoders.
\newblock In {\em NIPS Workshop on Bayesian Deep Learning}, 2016.

\bibitem{gui2022good}
Shurui Gui, Xiner Li, Limei Wang, and Shuiwang Ji.
\newblock Good: A graph out-of-distribution benchmark.
\newblock In {\em Advances in Neural Information Processing Systems 35 (NeurIPS 2022)}, 2022.
\newblock Datasets and Benchmarks Track.

\bibitem{zhao2024fug}
Jitao Zhao, Di~Jin, Meng Ge, Lianze Shan, Xin Wang, Dongxiao He, and Zhiyong Feng.
\newblock Fug: Feature-universal graph contrastive pre-training for graphs with diverse node features.
\newblock In {\em Proceedings of the 38th Conference on Neural Information Processing Systems (NeurIPS 2024)}, 2024.

\bibitem{zhao2019ppgn}
Cheng Zhao, Chenliang Li, and Cong Fu.
\newblock Cross-domain recommendation via preference propagation graphnet.
\newblock In {\em Proceedings of the 28th ACM International Conference on Information and Knowledge Management (CIKM 2019)}, pages 2165--2168, 2019.

\bibitem{zhao2020catn}
Cheng Zhao, Chenliang Li, Rong Xiao, Hongbo Deng, and Aixin Sun.
\newblock Catn: Cross-domain recommendation for cold-start users via aspect transfer network.
\newblock In {\em Proceedings of the 43rd International ACM SIGIR Conference on Research and Development in Information Retrieval (SIGIR 2020)}, pages 229--238, 2020.

\bibitem{zhu2021transfer}
Qi~Zhu, Carl Yang, Yidan Xu, Haonan Wang, Chao Zhang, and Jiawei Han.
\newblock Transfer learning of graph neural networks with ego-graph information maximization.
\newblock In {\em Advances in Neural Information Processing Systems 34 (NeurIPS 2021)}, 2021.

\bibitem{huang2025enhancing}
Xuanwen Huang, Wei Chow, Yize Zhu, Yang Wang, Ziwei Chai, Chunping Wang, Lei Chen, and Yang Yang.
\newblock Enhancing cross-domain link prediction via evolution process modeling.
\newblock In {\em Proceedings of the ACM Web Conference 2025 (WWW 2025)}, pages 2158--2171, 2025.

\bibitem{zhao2023glem}
Jianan Zhao, Meng Qu, Chaozhuo Li, Hao Yan, Qian Liu, Rui Li, Xing Xie, and Jian Tang.
\newblock Learning on large-scale text-attributed graphs via variational inference.
\newblock In {\em Proceedings of the 11th International Conference on Learning Representations (ICLR 2023)}, 2023.

\bibitem{yang2016planetoid}
Zhilin Yang, William~W. Cohen, and Ruslan Salakhutdinov.
\newblock Revisiting semi-supervised learning with graph embeddings.
\newblock In {\em Proceedings of the 33rd International Conference on Machine Learning (ICML 2016)}, 2016.

\bibitem{shi2023tradeoff}
Zhenmei Shi, Jiefeng Chen, Kunyang Li, Jayaram Raghuram, Xi~Wu, Yingyu Liang, and Somesh Jha.
\newblock The trade-off between universality and label efficiency of representations from contrastive learning.
\newblock In {\em Proceedings of the 11th International Conference on Learning Representations}, 2023.

\bibitem{mackiewicz1993pca}
Andrzej Mackiewicz and Waldemar Ratajczak.
\newblock Principal components analysis ({PCA}).
\newblock {\em Computers \& Geosciences}, 19(3):303--342, 1993.

\bibitem{kipf2017semi}
Thomas~N. Kipf and Max Welling.
\newblock Semi-supervised classification with graph convolutional networks.
\newblock In {\em Proceedings of the 5th International Conference on Learning Representations}, 2017.

\bibitem{romero2017diet}
Adriana Romero, Pierre~Luc Carrier, Akram Erraqabi, Tristan Sylvain, Alex Auvolat, Etienne Dejoie, Marc-Andr{\'e} Legault, Marie-Pierre Dub{\'e}, Julie~G. Hussin, and Yoshua Bengio.
\newblock Diet networks: Thin parameters for fat genomics.
\newblock In {\em Proceedings of the 5th International Conference on Learning Representations (ICLR 2017)}, 2017.

\bibitem{zhang2022localized}
Hengrui Zhang, Qitian Wu, Yu~Wang, Shaofeng Zhang, Junchi Yan, and Philip~S. Yu.
\newblock Localized contrastive learning on graphs.
\newblock {\em arXiv preprint arXiv:2212.04604}, 2022.

\bibitem{lee2022augmentation}
Namkyeong Lee, Junseok Lee, and Chanyoung Park.
\newblock Augmentation-free self-supervised learning on graphs.
\newblock In {\em Proceedings of the Thirty-Sixth AAAI Conference on Artificial Intelligence}, pages 7372--7380, 2022.

\bibitem{neal1998view}
Radford~M. Neal and Geoffrey~E. Hinton.
\newblock A view of the {EM} algorithm that justifies incremental, sparse, and other variants.
\newblock In {\em Learning in Graphical Models}, pages 355--368. Kluwer Academic Publishers, 1998.

\bibitem{besag1975statistical}
Julian Besag.
\newblock Statistical analysis of non-lattice data.
\newblock {\em The Statistician}, 24(3):179--195, 1975.

\bibitem{li2022ood}
Haoyang Li, Ziwei Zhang, Xin Wang, and Wenwu Zhu.
\newblock Learning invariant graph representations for out-of-distribution generalization.
\newblock In {\em Proceedings of the 36th Conference on Neural Information Processing Systems (NeurIPS 2022)}, 2022.

\bibitem{weinberger2009featurehashing}
Kilian Weinberger, Anirban Dasgupta, Josh Attenberg, John Langford, and Alex Smola.
\newblock Feature hashing for large scale multitask learning.
\newblock In {\em Proceedings of the 26th International Conference on Machine Learning (ICML 2009)}, 2009.

\bibitem{song2020mpnet}
Kaitao Song, Xu~Tan, Tao Qin, Jianfeng Lu, and Tie-Yan Liu.
\newblock Mpnet: Masked and permuted pre-training for language understanding.
\newblock In {\em Proceedings of the 34th Conference on Neural Information Processing Systems (NeurIPS 2020)}, 2020.

\bibitem{baltrusaitis2019multimodal}
Tadas Baltru{\v{s}}aitis, Chaitanya Ahuja, and Louis-Philippe Morency.
\newblock Multimodal machine learning: A survey and taxonomy.
\newblock {\em IEEE Transactions on Pattern Analysis and Machine Intelligence}, 41(2):423--443, 2019.

\bibitem{ni2019amazon}
Jianmo Ni, Jiacheng Li, and Julian McAuley.
\newblock Justifying recommendations using distantly-labeled reviews and fine-grained aspects.
\newblock In {\em Proceedings of the 2019 Conference on Empirical Methods in Natural Language Processing and the 9th International Joint Conference on Natural Language Processing (EMNLP-IJCNLP 2019)}, 2019.

\bibitem{mcauley2015image}
Julian McAuley, Christopher Targett, Qinfeng Shi, and Anton van~den Hengel.
\newblock Image-based recommendations on styles and substitutes.
\newblock In {\em Proceedings of the 38th International ACM SIGIR Conference on Research and Development in Information Retrieval (SIGIR 2015)}, 2015.

\bibitem{priem2022openalex}
Jason Priem, Heather Piwowar, and Richard Orr.
\newblock Openalex: A fully-open index of scholarly works, authors, venues, institutions, and concepts.
\newblock arXiv preprint arXiv:2205.01833, 2022.

\bibitem{chen2021simsiam}
Xinlei Chen and Kaiming He.
\newblock Exploring simple siamese representation learning.
\newblock In {\em Proceedings of the IEEE/CVF Conference on Computer Vision and Pattern Recognition (CVPR 2021)}, 2021.

\bibitem{sanh2019distilbert}
Victor Sanh, Lysandre Debut, Julien Chaumond, and Thomas Wolf.
\newblock Distilbert, a distilled version of bert: Smaller, faster, cheaper and lighter.
\newblock arXiv preprint arXiv:1910.01108, 2019.

\bibitem{grill2020byol}
Jean-Bastien Grill, Florian Strub, Florent Altch{\'e}, Corentin Tallec, Pierre~H. Richemond, Elena Buchatskaya, Carl Doersch, Bernardo~Avila Pires, Zhaohan~Daniel Guo, Mohammad~Gheshlaghi Azar, Bilal Piot, Koray Kavukcuoglu, R{\'e}mi Munos, and Michal Valko.
\newblock Bootstrap your own latent: A new approach to self-supervised learning.
\newblock In {\em Advances in Neural Information Processing Systems (NeurIPS 2020)}, 2020.

\bibitem{reimers2019sbert}
Nils Reimers and Iryna Gurevych.
\newblock Sentence-bert: Sentence embeddings using siamese bert-networks.
\newblock In {\em Proceedings of the 2019 Conference on Empirical Methods in Natural Language Processing and the 9th International Joint Conference on Natural Language Processing (EMNLP-IJCNLP 2019)}, 2019.

\bibitem{hinton2015distilling}
Geoffrey Hinton, Oriol Vinyals, and Jeff Dean.
\newblock Distilling the knowledge in a neural network.
\newblock {\em arXiv preprint arXiv:1503.02531}, 2015.

\bibitem{romero2015fitnets}
Adriana Romero, Nicolas Ballas, Samira Ebrahimi~Kahou, Antoine Chassang, Carlo Gatta, and Yoshua Bengio.
\newblock Fitnets: Hints for thin deep nets.
\newblock In {\em Proceedings of the 3rd International Conference on Learning Representations (ICLR 2015)}, 2015.

\end{thebibliography}

\appendix

\section{Results of the Additional Experiments}

\begin{table}[H]
\centering
\caption{Comparison of cross-domain transfer performance across experimental settings.}
\label{tab:cross_domain_results}
\begin{threeparttable}
\small
\setlength{\tabcolsep}{3.5pt}
\begin{tabularx}{\textwidth}{lXcccc}
\toprule
Experiment & Method & \makecell{Full GCN \\ ($Z$)} & \makecell{Raw Text \\ Hash} & \makecell{Raw \\ MPNet} & Ensemble \\
\midrule
Exp1 & FUG-only & 0.7459 & 0.7623 & -- & 0.7702 \\
Exp2 & GCN pseudo-($Z$) imitation TextHead & 0.7445 & 0.7623 & -- & 0.7695 \\
Exp3 & Raw text-hash external anchor & 0.7437 & 0.7623 & -- & 0.7717 \\
Exp4 & Frozen MPNet external teacher & 0.7437 & 0.7623 & 0.7888 & 0.7724 / 0.7845\tnote{a} \\
\bottomrule
\end{tabularx}
\begin{tablenotes}
\footnotesize
\item[a] The value $0.7845$ denotes the GCN+MPNet ensemble; the value $0.7724$ denotes the GCN+Raw-Text-Hash ensemble.
\end{tablenotes}
\end{threeparttable}
\end{table}

For Exp2, checkpoint selection based on source validation balanced accuracy selects the pre-update warm-up checkpoint rather than any of the em1--em6 checkpoints after EM-like updates. The OpenAlex value of $0.7445$ in Table~\ref{tab:cross_domain_results} therefore evaluates the FUG representation after the 60-epoch warm-up, not a representation after self-copy distillation. The result shows that the post-distillation candidates did not improve the source-side selection criterion; it does not imply that the target performance of a post-distillation model was $0.7445$.

\begin{table}[H]
\centering
\caption{Test accuracy at different representation stages under the OpenAlex linear probe.}
\label{tab:openalex_representation_comparison}
\small
\setlength{\tabcolsep}{6pt}
\begin{tabular}{lcccc}
\toprule
Representation or difference
& Exp2
& Exp3
& Exp4
& FUG+GLEM-ITT \\
\midrule
\multicolumn{5}{l}{\textit{Test accuracy of each representation}} \\
\addlinespace[2pt]
Raw Text Hash
& $0.7623$
& $0.7623$
& $0.7623$
& $0.7623$ \\

MLP-only (before GCN propagation)
& $0.7380$
& $0.7380$
& $0.7373$
& $0.7359$ \\

Full GCN $Z$ (after GCN propagation)
& $0.7445$
& $0.7437$
& $0.7437$
& $0.7480$ \\

Raw MPNet
& --
& --
& $0.7888$
& -- \\

Ensemble (GCN $Z$ + Text Hash)
& $0.7695$
& $0.7717$
& $0.7724$
& $0.7717$ \\

Ensemble (GCN $Z$ + MPNet)
& --
& --
& $0.7845$
& -- \\
\midrule
\multicolumn{5}{l}{\textit{Difference between Full GCN $Z$ and each representation}} \\
\addlinespace[2pt]
$\Delta_{\text{Hash}\rightarrow\text{GCN}}
=
\text{Acc}(Z_{\text{GCN}})
-
\text{Acc}(\text{Raw Text Hash})$
& $-0.0178$
& $-0.0186$
& $-0.0186$
& $-0.0143$ \\

$\Delta_{\text{MLP}\rightarrow\text{GCN}}
=
\text{Acc}(Z_{\text{GCN}})
-
\text{Acc}(\text{MLP-only})$
& $+0.0065$
& $+0.0057$
& $+0.0064$
& $+0.0121$ \\

$\Delta_{\text{MPNet}\rightarrow\text{GCN}}
=
\text{Acc}(Z_{\text{GCN}})
-
\text{Acc}(\text{Raw MPNet})$
& --
& --
& $-0.0451$
& -- \\
\bottomrule
\end{tabular}

\vspace{3pt}
\begin{minipage}{0.97\linewidth}
\footnotesize
All entries are test accuracies from the standard linear probe on OpenAlex. Negative values of $\Delta_{\text{Hash}\rightarrow\text{GCN}}$ and $\Delta_{\text{MPNet}\rightarrow\text{GCN}}$ indicate that Full GCN $Z$ underperforms the corresponding raw-text representation. Positive values of $\Delta_{\text{MLP}\rightarrow\text{GCN}}$ indicate that GCN propagation provides some improvement over the MLP-only representation.
\end{minipage}
\end{table}

\FloatBarrier

\begin{table}[H]
\centering
\caption{Representation-level classification performance of FUG-only and FUG+GLEM-ITT on OpenAlex.}
\label{tab:representation_performance}
\small
\setlength{\tabcolsep}{6pt}
\begin{tabular}{lcccc}
\toprule
Method
& \makecell{Raw Text\\Hash}
& MLP-only
& \makecell{Full GCN\\($Z$)}
& Ensemble \\
\midrule
FUG-only
& $0.7623$
& $0.7480$
& $0.7459$
& $0.7702$ \\

FUG+GLEM-ITT
& $0.7623$
& $0.7359$
& $0.7480$
& $0.7717$ \\
\bottomrule
\end{tabular}

\vspace{2pt}
\begin{minipage}{0.95\linewidth}
\footnotesize
All entries are test accuracies from the standard linear probe on OpenAlex. MLP-only is the internal FUG representation before GCN propagation, Full GCN ($Z$) is the post-propagation node representation, and Ensemble is the concatenation of Full GCN ($Z$) and Raw Text Hash.
\end{minipage}
\end{table}

\begin{table}[H]
\centering
\caption{Changes in alignment metrics and Full GCN-$Z$ classification performance on OpenAlex.}
\label{tab:alignment_results}
\small
\setlength{\tabcolsep}{5pt}
\begin{tabular}{llcccccc}
\toprule
Setting
& Metric
& Direction
& Initial
& Final
& Change
& \makecell{Full GCN $Z$\\test accuracy}
& \makecell{Difference from\\FUG-only} \\
\midrule

\multicolumn{8}{l}{
\textit{(a) Cosine similarity: larger values indicate stronger anchor alignment}
} \\
\addlinespace[2pt]

Exp3
& \texttt{ood\_cos}
& $\uparrow$
& $0.0003$
& $0.6918$
& $+0.6915$
& $0.7437$
& $-0.0022$ \\

Exp3
& \texttt{raw\_cos}
& $\uparrow$
& $-0.0084$
& $0.2693$
& $+0.2777$
& --
& -- \\

Exp4
& \texttt{mpnet\_cos}
& $\uparrow$
& $0.0095$
& $0.4341$
& $+0.4246$
& $0.7437$
& $-0.0022$ \\

\midrule

\multicolumn{8}{l}{
\textit{(b) Cosine anchor loss: smaller values indicate stronger anchor alignment}
} \\
\addlinespace[2pt]

FUG+GLEM-ITT
& \texttt{text\_anc}
& $\downarrow$
& $0.8032$
& $0.3484$
& $-0.4548$
& $0.7480$
& $+0.0021$ \\

FUG+GLEM-ITT
& \texttt{mlp\_anc}
& $\downarrow$
& $0.8712$
& $0.5157$
& $-0.3555$
& --
& -- \\

FUG+GLEM-ITT
& \texttt{raw\_anc}
& $\downarrow$
& $0.8836$
& $0.6132$
& $-0.2704$
& --
& -- \\

\bottomrule
\end{tabular}

\vspace{3pt}
\begin{minipage}{0.97\linewidth}
\footnotesize
Full GCN $Z$ is evaluated by test accuracy under the standard linear probe on OpenAlex. The FUG-only Full GCN-$Z$ accuracy is $0.7459$; ``Difference from FUG-only'' subtracts this value from each model's Full GCN-$Z$ accuracy. For Exp3 and Exp4, Initial denotes the end of warm-up and Final denotes the selected final candidate (Iter~6 in both cases). For the FUG+GLEM-ITT anchor losses, Initial is epoch~6 of the Iter~1 M-step and Final is epoch~18 of the Iter~6 M-step(The M-step consists of 18 epochs per iteration.). Because \texttt{text\_anc}, \texttt{mlp\_anc}, and \texttt{raw\_anc} are losses of the form $1-\cos(\mathbf{Z}_{\text{GCN}},\mathbf{A})$, a decrease indicates stronger alignment between $\mathbf{Z}_{\text{GCN}}$ and the anchor representation.
\end{minipage}
\end{table}

\begin{table}[H]
\centering
\caption{Candidate-model metrics and the FUG+GLEM-ITT model-selection result.}
\label{tab:rev90_candidate_selection}
\small
\setlength{\tabcolsep}{5pt}
\begin{tabular}{lcccccc}
\toprule
Candidate
& \code{valid\_bacc}
& \code{ood\_cos}
& \code{text\_cos}
& Composite
& \makecell{Difference from Warmup\\(\code{valid\_bacc})}
& Selected \\
\midrule
Warmup
& $0.4788$
& $0.0003$
& $-0.0166$
& $0.1132$
& $0.0000$
& -- \\

em1
& $0.4707$
& $0.2826$
& $0.4404$
& $0.3927$
& $-0.0081$
& -- \\

em2
& $0.4715$
& $0.3967$
& $0.5576$
& $0.4797$
& $-0.0073$
& -- \\

em3
& $0.4765$
& $0.4554$
& $0.6097$
& $0.5224$
& $-0.0023$
& -- \\

em4
& $0.4723$
& $0.4742$
& $0.6328$
& $0.5372$
& $-0.0065$
& -- \\

em5
& $0.4746$
& $0.4807$
& $0.6444$
& $0.5447$
& $-0.0042$
& -- \\

em6
& $0.4766$
& $0.4845$
& $0.6512$
& $\mathbf{0.5492}$
& $-0.0022$
& $\checkmark$ \\

SWA
& $\mathbf{0.4789}$
& $0.4268$
& $0.6135$
& $0.5145$
& $+0.0001$
& -- \\
\bottomrule
\end{tabular}

\vspace{3pt}
\begin{minipage}{0.97\linewidth}
\footnotesize
The composite score is $0.25\times\texttt{valid\_bacc}+0.35\times\texttt{ood\_cos}+0.40\times\texttt{text\_cos}$. Boldface marks the maximum composite score and the maximum \code{valid\_bacc}, respectively. Because model selection uses the composite score, em6 is selected instead of SWA, even though SWA has the highest source-side \code{valid\_bacc}. The em6 checkpoint is obtained after the E-like refresh in the sixth EM-like iteration and the subsequent 18-epoch M-step.
\end{minipage}
\end{table}

\section{Implementation and Evaluation Details}

\paragraph{Loss weights for FUG+GLEM-ITT.}

The FUG+GLEM-ITT loss weights are
\begin{equation}
\label{eq:appendix_weights}
W_{\mathrm{FUG}}=0.80, \quad W_{\mathrm{text}}=1.80, \quad W_{\mathrm{MLP}}=1.10, \quad W_{\mathrm{raw}}=0.70, \quad W_{\mathrm{ext}}=0.25.
\end{equation}

\paragraph{Data construction and splitting.}

For Digital Music, graph nodes are matched to reviews by \texttt{(user\_id, asin, timestamp)}. Each review is treated as a node, the concatenation of its title and body is used as the node text, and its rating is mapped to one of five classes from 0 to 4. Edges between matched nodes are converted to undirected edges.

For OpenAlex, each paper record is treated as a node. Node text is selected in the priority order \texttt{matched title}, \texttt{\_source\_title}, and \texttt{text}. When concepts are available, the highest-scoring concept is used as the label; otherwise, document type is used. Only the 20 most frequent classes are retained. Edges between paper nodes are converted to undirected edges.

\paragraph{Text features.}

Each node's text is represented by concatenating word-level and character-level hashing features. Word unigrams and bigrams are hashed into 1,024 dimensions. Character $n$-grams with $n=3,\ldots,5$ are hashed into a separate 1,024-dimensional vector. The resulting graph-model input has 2,048 text dimensions and is $\ell_2$-normalized after concatenation.

\paragraph{Source-domain split.}

The 130,434 matched Digital Music nodes are split according to 128,764 unique \texttt{(user\_id, asin, timestamp)} metadata tuples. A total of 1,670 nodes share a metadata tuple with another node. The metadata tuples are divided into train, validation, and test sets in a $60{:}20{:}20$ ratio, yielding 77,258, 25,753, and 25,753 tuples, respectively. All nodes sharing a tuple are assigned to the same split, preventing leakage across splits due to duplicate reviews.

\paragraph{Target-domain split and use in evaluation.}

For OpenAlex, we scan the first 20,000 records in the JSONL file and retain records with text and labels that belong to one of the 20 most frequent classes. This process yields a target graph with 6,984 nodes and 20 classes. For linear-probe evaluation, the nodes are divided into train, validation, and test sets in a $60{:}20{:}20$ ratio, yielding 4,190, 1,397, and 1,397 nodes. Target labels are used only to train, validate, and evaluate the linear probe; they are not used to train the source model or select its checkpoint.

\paragraph{Target-feature standardization.}

Before constructing the linear-probe split, each dimension of the 2,048-dimensional OpenAlex text-hash features is standardized using the mean and standard deviation over all 6,984 retained nodes. This transformation does not use target labels, but it does access unlabeled features from the entire target domain that are available at evaluation time. The setting therefore includes label-free transductive preprocessing.

\paragraph{Fallback graph construction.}

For each domain, retained original edges are converted to undirected edges. The experiments retain 547,494 Digital Music edges and 3,830 OpenAlex edges, so the original edges are used in both cases. The implementation includes a fallback that constructs a $k=10$ nearest-neighbor graph from the 2,048-dimensional text-hash features only when no retained edge is available; this fallback is not used in the reported runs.

\begin{table}[H]
\centering
\caption{Checkpoint-selection procedures for FUG+GLEM-ITT and Exp1--Exp4.}
\label{tab:checkpoint_selection}
\begin{threeparttable}
\footnotesize
\setlength{\tabcolsep}{3.5pt}
\begin{tabularx}{\textwidth}{l l X X}
\toprule
Experiment & Candidate checkpoints & Selection score & Notes \\
\midrule
FUG+GLEM-ITT &
Warmup, EM1--EM6, SWA &
$
S =
0.25Bal_{\mathrm{val}}
+
0.35Cos_{\mathrm{MLP}}
+
0.40Cos_{\mathrm{TextHead}}
$
&
Warmup, the end of each EM iteration, and the SWA model obtained by averaging model states after the EM iterations are candidates. The final checkpoint maximizes a composite of source-domain validation balanced accuracy and the mean cosine similarities, computed over all source-graph nodes, between Full GCN $Z$ and the MLP-only and TextHead representations. \\
\addlinespace
Exp1 &
After epochs $10,20,\ldots,160,168$ &
$S=Bal_{\mathrm{val}}$
&
Selects the FUG-only checkpoint with the highest source validation balanced accuracy. \\
\addlinespace
Exp2 &
Warmup, EM1--EM6 &
$S=Bal_{\mathrm{val}}$
&
The cosine similarity between GCN $Z$ and the TextHead output is recorded but not used for selection. \\
\addlinespace
Exp3 &
Warmup, EM1--EM6 &
$
S_{\mathrm{impl}} =
0.35Bal_{\mathrm{val}}
+
0.30Cos_{\mathrm{MLP}}
$
&
Cosine similarity to the raw text-hash anchor is recorded, but it is not passed to the shared selection function as \texttt{text\_cos} and is therefore absent from the selection score. \\
\addlinespace
Exp4 &
Warmup, EM1--EM6 &
$
S =
0.35Bal_{\mathrm{val}}
+
0.30Cos_{\mathrm{MLP}}
+
0.35Cos_{\mathrm{MPNet}}
$
&
The cosine similarity between GCN $Z$ and the frozen MPNet anchor is passed as \texttt{text\_cos} and included in the selection score. \\
\bottomrule
\end{tabularx}
\begin{tablenotes}
\footnotesize
\item $Bal_{\mathrm{val}}$ is balanced accuracy on the source-domain validation split.
\item $Cos_{\mathrm{MLP}}$ is the mean cosine similarity between Full GCN $Z$ and the MLP-only representation.
\item $Cos_{\mathrm{TextHead}}$ and $Cos_{\mathrm{MPNet}}$ are the mean cosine similarities between Full GCN $Z$ and the TextHead representation and frozen MPNet anchor, respectively.
\item OpenAlex labels and target-domain evaluation results are not used for checkpoint selection in any setting.
\end{tablenotes}
\end{threeparttable}
\end{table}

\section{Details of Findings}

\subsection{Detailed Analysis of Text-to-GCN Projection}
\label{app:text_to_gcn_projection}

In Exp4, MPNet embeddings are not fed directly into the GCN. They are projected into \code{anchor\_mpnet}, which has the same dimensionality as GCN $Z$. Specifically, the MPNet representation is transformed from \code{raw\_dim}$=768$ to a \code{projected\_dim}$=1024$ anchor. The cosine loss is applied between GCN $Z$ and this projected MPNet anchor, rather than between GCN $Z$ and the original MPNet representation. Some of the semantic structure encoded by MPNet may therefore be lost at this stage. Moreover, the MPNet space is not guaranteed to have explicitly learned directions that separate every OpenAlex class, such as reinforcement learning, adversarial learning, convolutional neural networks, and combinatorics. Projecting it into the GCN-$Z$ space need not preserve the fine-grained directions useful for OpenAlex classification.

The same issue applies to Raw Text Hash. We concatenate 1,024 word-$n$-gram dimensions and 1,024 character-$n$-gram dimensions to obtain a 2,048-dimensional text representation. In contrast, the hidden representation in FUG+GLEM-ITT, and hence GCN $Z$, has 1,024 dimensions. The lexical and character-$n$-gram information in Raw Text Hash is therefore compressed and projected into a 1,024-dimensional GCN-$Z$ space. This compression may contribute to the decrease from an accuracy of $0.7623$ for Raw Text Hash to approximately $0.7459$--$0.7480$ for Full GCN $Z$. The teacher knowledge is not transferred wholesale as a semantic space; instead, it is supplied as a target vector toward which the representation is moved by a cosine loss. The use of a projector to align intermediate representations of different dimensionalities is structurally related to FitNets, which transfers a teacher's intermediate representation to a student as a hint~\cite{romero2015fitnets}. In both FitNets and our setting, however, aligning representation spaces does not guarantee that directions useful for target classification survive compression and projection. The subsequent GCN propagation further means that proximity to the teacher representation need not improve classification. Teacher semantics are thus introduced into the GCN as a directional constraint, rather than as a classification-effective structure, producing transformation loss in addition to information compression.

\subsection{Mismatch between the Objectives of the Teacher and FUG Representation Spaces}
\label{app:Mismatch between the Objectives of the Teacher and FUG Representation Spaces}

The FUG self-supervised objective consists of multiple loss terms with weights $\text{\code{lambda\_neg}}=1.0$, $\text{\code{lambda\_dim}}=200.0$, and $\text{\code{lambda\_pos}}=0.5$. However, because these loss terms may differ in their numerical ranges and empirical scales during training, their relative dominance cannot be inferred from the coefficient values alone. Therefore, based solely on $\text{\code{lambda\_dim}}=200.0$, we do not claim that the dimension loss dominates the optimization or that it constrains GCN $Z$ to preserve the original FUG self-supervised geometry.

Nevertheless, the teacher anchor does not replace the FUG self-supervised objectives; rather, it constitutes one of several training signals that are optimized jointly. Consequently, the final GCN $Z$ does not reproduce the teacher space alone, but instead represents a compromise among alignment with the teacher, the FUG self-supervised objectives, GCN propagation, and the structure of the source graph. Thus, even if the distance between GCN $Z$ and the teacher decreases, directions in the teacher space that are useful for OpenAlex classification are not necessarily incorporated selectively into GCN $Z$.

In Exp4, \code{mpnet\_cos}, which measures alignment with MPNet, increased from $0.0095$ after warm-up to $0.4341$ at em6. This result confirms that the optimization successfully moved GCN $Z$ closer to the MPNet anchor. Nevertheless, the OpenAlex balanced accuracy of Full GCN $Z$ remained at $0.7437$. This finding indicates that bringing GCN $Z$ closer to MPNet does not necessarily align it in directions that are useful for the OpenAlex classification boundary. However, this result alone does not identify the contributions of the individual self-supervised loss terms or establish that any specific loss term interfered with the effect of the teacher anchor.

\subsection{Comparison of Balanced Accuracy Before and After GCN Propagation}
\label{app:Comparison of Balanced Accuracy}

On the OpenAlex test set, the balanced accuracies in Exp4 were $0.7623$ for Raw Text Hash and $0.7888$ for Raw MPNet, compared with $0.7437$ for Full GCN $Z$ (see the Exp4 column of Table~\ref{tab:openalex_representation_comparison}). This comparison indicates that the node-level text representations contain information useful for target classification, whereas the representation obtained after GCN propagation did not achieve comparable performance.

The FUG+GLEM-ITT-related experiments exhibited a similar pattern. In the FUG-only setting, the balanced accuracies were $0.7623$ for Raw Text Hash, $0.7480$ for MLP-only, $0.7459$ for Full GCN $Z$, and $0.7702$ for the ensemble. Thus, Full GCN $Z$ not only underperformed Raw Text Hash but also showed a slight decrease relative to the pre-propagation MLP-only representation. Similarly, in FUG+GLEM-ITT with TextHead used as an external anchor, Full GCN $Z$ achieved $0.7480$, which was lower than both Raw Text Hash at $0.7623$ and the ensemble at $0.7717$ (Table~\ref{tab:openalex_representation_comparison}). These results indicate that, at least in terms of balanced accuracy, the improvement obtained through GCN propagation was limited and that propagation occasionally reduced performance relative to the pre-propagation representation.

Strictly speaking, a GCN does not compute a simple arithmetic average of node representations. Instead, it aggregates a node's own representation and those of its neighbors using a normalized adjacency structure and learned transformations. When edges tend to connect nodes with the same label, such neighborhood aggregation can be effective. In the present cross-domain setting, however, the source domain is Amazon Music and the target domain is OpenAlex. The product- and user-behavior-related structure encoded by Amazon Music differs from the paper, concept, and citation relationships represented in OpenAlex. Therefore, the inductive bias toward neighborhood smoothing learned from the source graph may not transfer directly to OpenAlex concept classification. As a result, node-specific semantic information may be mixed with neighborhood information that is not useful for target classification, preventing the discriminative information derived from text from being fully exploited.

Nevertheless, the performance ordering described above is based on balanced accuracy and need not hold consistently for other evaluation metrics. Moreover, although these results reveal performance differences between representations before and after GCN propagation, they do not directly demonstrate that textual information is diluted by propagation. Stronger mechanistic evidence would require additional measurements of class separability before and after GCN propagation, the neighborhood label-mixing rate, the degree of homophily in the target graph, and the relative contributions of self-representations and neighborhood representations.

\subsection{Trade-off Between FUG SSL and Teacher Alignment}
\label{app:Trade-off Between FUG SSL and Teacher Alignment}

The M-step does not optimize the teacher anchor alone; the FUG SSL loss remains active during optimization. In Exp4, the M-step is constructed as a combination of FUG SSL and a frozen MPNet semantic anchor. Thus, during training, GCN $Z$ is simultaneously influenced by the FUG SSL objective, which seeks to preserve a stable self-supervised representation on the source graph, and the MPNet anchor objective, which encourages alignment with textual semantics. These two objectives do not necessarily favor the same representational geometry. Alignment with the source graph tends to organize representations according to the review and purchasing relationships in Amazon Music, whereas alignment with MPNet tends to organize them according to textual semantics. In contrast, the OpenAlex target task requires a representation that is useful for academic concept classification. Because these three structures need not coincide, the M-step may converge to a compromise solution. In such a solution, alignment with MPNet may remain limited in order to preserve the FUG geometry, or stronger alignment with MPNet may alter the FUG geometry without producing a sufficient improvement in OpenAlex classification performance.

Table~\ref{tab:rev90_candidate_selection} reports the source validation performance, alignment metrics, and composite scores of the candidate models in FUG+GLEM-ITT. The \code{valid\_bacc} at warmup was $0.4788$, whereas the value for the ultimately selected em6 model---the FUG encoder candidate obtained after the sixth EM-like iteration---was $0.4766$. Although em6 had a slightly lower source validation balanced accuracy than warmup, it achieved the highest composite score, defined as
\begin{equation}
0.25 \times \code{valid\_bacc}
+0.35 \times \code{ood\_cos}
+0.40 \times \code{text\_cos}.
\end{equation}

By contrast, SWA achieved a relatively high source-side score of \code{valid\_bacc}$=0.4789$
but had a lower composite score than em6 and was therefore not selected. This result indicates that preserving source validation performance and approaching the text anchor are not fully aligned under the current selection criterion.

For model selection, we saved warmup, the models obtained after each EM-like iteration from em1 to em6, and SWA. We computed the composite score above for each candidate and selected the candidate with the highest value as the final model. In this criterion, source \code{valid\_bacc} is assigned a weight of $0.25$, \code{ood\_cos},
which measures alignment between Full GCN $Z$ and the MLP-only representation, is assigned a weight of $0.35$, and \code{text\_cos}, which measures alignment with the TextHead anchor, is assigned a weight of $0.40$. Thus, the two alignment metrics receive a larger combined nominal weight than source validation performance. Consequently, a candidate obtained after an EM-like update may be selected even when its source \code{valid\_bacc} is lower than that of warmup. This observation further indicates that preserving source validation performance and aligning the Full GCN representation with the MLP-only representation and TextHead anchor are not necessarily identical objectives.

In the M-step of FUG+GLEM-ITT, the FUG SSL objective is retained with $W_{\mathrm{FUG}}=0.80$, while the TextHead anchor, MLP-only anchor, raw-hash anchor, and auxiliary random-projection text anchor are assigned weights of $W_{\mathrm{text}}=1.80$, $W_{\mathrm{MLP}}=1.10$, $W_{\mathrm{raw}}=0.70$, and $W_{\mathrm{ext}}=0.25$, respectively. Because these loss terms may differ in their numerical and gradient scales, the coefficient values alone do not establish their effective relative influence. Moreover, these weights were set empirically based on preliminary experiments rather than obtained through an exhaustive hyperparameter search. Accordingly, this configuration should not be interpreted as a universally optimal setting for maximizing target accuracy on OpenAlex, but rather as a representative setting for analyzing the trade-off between FUG SSL and multiple anchors.

Specifically, rather than removing FUG SSL and aligning the representation exclusively with the TextHead teacher, the proposed design retains FUG SSL and applies cosine-alignment losses to the post-propagation representation $Z_{\mathrm{GCN}}$. These losses use an independent TextHead anchor, a stop-gradient MLP-only anchor, and a fixed-projection raw text-hash anchor. An additional alignment loss with an external text anchor is also applied in an auxiliary random-projection space. Consequently, $Z_{\mathrm{GCN}}$ is updated as a compromise representation between the FUG self-supervised objective and the alignment objectives associated with multiple anchors. Taken together, the experimental results are consistent with a trade-off in which weak teacher influence provides only limited benefits, whereas stronger teacher alignment may alter the FUG geometry or reduce source validation performance. This trade-off may therefore be an important factor constraining improvements in target-domain performance.

\end{document}